\documentclass[11pt]{article}
\usepackage[preprint]{acl}

\usepackage{latexsym}
\usepackage{booktabs}
\usepackage{multirow}
\usepackage{array}
\usepackage{amsmath}
\usepackage{amssymb}
\usepackage{graphicx}
\usepackage{caption}
\usepackage{subcaption}
\usepackage{enumitem}
\usepackage{microtype}
\usepackage{xspace}
\usepackage{url}

\usepackage{xcolor}
\usepackage[most]{tcolorbox}
\usepackage{fvextra}

\usepackage{fontspec}
\usepackage{xeCJK}

\usepackage{tabularx}
\usepackage{makecell}

\setCJKsansfont{FandolHei-Regular.otf}[BoldFont=FandolHei-Bold.otf]
\setCJKmonofont{FandolFang-Regular.otf}

\newcommand{\method}{Wenyan Thinking\xspace}
\newcommand{\ours}{Wenyan (Qwen3.5 normal)\xspace}
\newcommand{\think}{Qwen3-4B-Thinking-2507\xspace}
\newcommand{\vern}{Long Modern-Chinese SFT\xspace}
\newcommand{\compact}{Compact Modern-Chinese\xspace}
\newcommand{\concise}{Concise-Prompt Think\xspace}

\tcbset{
  promptbox/.style={
    enhanced,
    breakable,
    sharp corners,
    boxrule=0.7pt,
    colback=white,
    colframe=black!70,
    left=2mm,
    right=2mm,
    top=1.5mm,
    bottom=1.5mm
  },
  promptzh/.style={
    promptbox,
    colback=blue!2,
    colframe=blue!50!black,
    fonttitle=\bfseries\small
  },
  prompten/.style={
    promptbox,
    colback=green!2,
    colframe=green!45!black,
    fonttitle=\bfseries\small
  }
}

\tcbset{
  appbox/.style={
    enhanced,
    breakable,
    sharp corners,
    boxrule=0.7pt,
    left=2mm,
    right=2mm,
    top=1.2mm,
    bottom=1.2mm,
    colback=white,
    colframe=black!70
  },
  querybox/.style={
    appbox,
    colback=gray!3,
    colframe=black!70,
    fonttitle=\bfseries
  },
  zhbox/.style={
    appbox,
    colback=blue!2,
    colframe=blue!55!black,
    fonttitle=\bfseries
  },
  enbox/.style={
    appbox,
    colback=green!2,
    colframe=green!45!black,
    fonttitle=\bfseries
  }
}

\title{Does the Linguistic Register of Thinking Matter? Wenyan as a Compact Reasoning Register for Chinese LLMs}

\author{Minghan Li
  \and
  Guodong Zhou \\
  Soochow University, Suzhou, China \\
  \texttt{mhli@suda.edu.cn, gdzhou@suda.edu.cn}}

\begin{document}
\maketitle

\begin{abstract}
Visible reasoning traces of Chinese reasoning-oriented LLMs are typically written in modern vernacular Chinese, which may be verbose under limited decoding budgets. We ask whether the linguistic register of reasoning can be used to improve efficiency. We propose \method, which emits intermediate reasoning in Classical Chinese (wenyanwen) while retaining a short direct-answer output. Starting from modern-Chinese reasoning traces, we rewrite only the reasoning field into wenyan and fine-tune \think on the rewritten supervision. To distinguish register shift from generic compression, we construct a compact modern-Chinese control using the same source data, teacher family, base model, and SFT recipe. Across budget-sensitive evaluations on  MMLU-ProX-Lite-ZH, CEVAL, and AGIEval-CN, \method substantially reduces reasoning length and missing-answer rates under constrained decoding. Compact rewriting explains much of the efficiency gain, while wenyan consistently outperforms the compact modern-Chinese control on the primary benchmark and remains competitive across the additional benchmarks. These results identify linguistic register as a practical design axis for efficient visible reasoning in Chinese LLMs.
\end{abstract}

\section{Introduction}

Visible reasoning traces have become a common interface for modern large language models (LLMs) \cite{wei2022chain,kojima2022large,wang2023selfconsistency,zelikman2022star,lightman2023let,guo2025deepseek,yang2025qwen3}. In reasoning-oriented systems, the model often emits a long chain of intermediate thoughts before producing the final answer. Such traces can improve performance on difficult problems, but they also consume tokens, latency, and user-visible generation budget. Under autoregressive decoding, a longer visible trace directly increases inference cost and raises the risk that the final answer is delayed, truncated, or formatted incorrectly. This cost side of reasoning is increasingly salient as recent work treats inference-time reasoning budget as a first-class design dimension \cite{snell2024scaling,han2024token,xu2025chain}.

For Chinese thinking models, these visible traces are usually written in modern vernacular Chinese. This is natural for readability, but it also raises a question that has received little direct attention: \emph{does the language register of the reasoning trace affect reasoning efficiency?} In particular, could broadly similar reasoning content be expressed more compactly if the model reasons in Classical Chinese (wenyanwen) rather than in modern Chinese?

This paper studies that question through \method. We keep the answer channel in the benchmark-required direct-answer format, but rewrite the visible reasoning channel into wenyan. The motivation is not literary nostalgia. Rather, Classical Chinese is historically compact: relations and discourse elements that are explicitly spelled out in modern Chinese can often be expressed more tersely. We therefore treat wenyan as a candidate high-density register for the visible reasoning channel of Chinese LLMs.

Our core contribution is focused. We do not claim to propose a general-purpose reasoning-compression framework, nor do we center the paper on teacher ensembling or prompt-style control. Instead, we ask a specific, under-explored question: can the visible reasoning traces of Chinese LLMs be shortened by shifting them from modern Chinese to wenyan, while keeping final answers in a short direct-answer format and largely preserving task performance?

To answer this question, we construct \textsc{WenyanThinking-10k} by rewriting modern-Chinese reasoning traces into wenyan with strong teacher models and then fine-tuning \think so that it learns the format
\begin{equation*}
\begin{split}
&\texttt{<think> wenyan reasoning </think>} \\
&\texttt{direct answer}.
\end{split}
\end{equation*}
This design preserves answer readability while explicitly changing the linguistic register of the reasoning channel. Importantly, we also construct a \compact control from the same source examples, teacher family, and downstream SFT recipe. This control lets us separate generic compact rewriting from the specific effect of shifting the reasoning register to wenyan.

The paper is organized around a single main story: whether a register shift in the visible reasoning channel improves the accuracy--budget--latency trade-off. Prompt-style and teacher-source analyses are reported as appendices to clarify robustness without redefining the central claim.

Our contributions are as follows:
\begin{enumerate}[leftmargin=*,itemsep=2pt]
    \item We propose \method, a setting in which a Chinese thinking model performs visible intermediate reasoning in Classical Chinese while preserving a short, benchmark-compatible direct-answer output format.
    \item We construct a practical wenyan-rewritten supervision set, \textsc{WenyanThinking-10k}, from modern-Chinese reasoning traces, and fine-tune a Chinese thinking model on this supervision without modifying the model architecture.
    \item We introduce a compact modern-Chinese control that uses the same source examples, teacher family, base model, and training pipeline but rewrites reasoning traces into concise modern Chinese, enabling us to disentangle generic compact rewriting from register shift.
    \item Through budget-sensitive evaluations and multiple baselines, we show that wenyan supervision yields a favorable accuracy--budget--latency trade-off under constrained decoding.
\end{enumerate}

\section{Related Work}

\paragraph{Visible reasoning and chain-of-thought.}
Chain-of-thought (CoT) prompting showed that exposing intermediate reasoning substantially improves performance on challenging problems \cite{wei2022chain}, including mathematical and symbolic settings that were later popularized through GSM8K and BIG-Bench Hard style evaluations \cite{cobbe2021training,suzgun2022challenging}. Subsequent work extended this direction through zero-shot CoT \cite{kojima2022large}, self-consistency decoding \cite{wang2023selfconsistency}, least-to-most decomposition \cite{zhou2023least}, self-generated rationales \cite{zelikman2022star}, tool- or program-mediated reasoning \cite{yao2023react,gao2023pal,chen2023program}, tree-structured search over thoughts \cite{yao2023tree}, process reward models \cite{lightman2023let}, and large-scale reasoning-oriented post-training \cite{guo2025deepseek,yang2025qwen3}. These efforts established that \emph{what} reasoning content a model generates matters greatly, but they largely treat the linguistic form of reasoning traces as a fixed background condition. Our work departs from this default by asking whether the \emph{language register} of the reasoning channel itself is a meaningful design variable.

\paragraph{Reasoning efficiency and token economy.}
The token inefficiency of long CoT traces has drawn increasing attention. Recent work explores concise rationale generation through explicit length control, token-budget-aware training, conditioned CoT compression, draft-style reasoning, summary-style distillation, and self-training for shorter reasoning traces \cite{fu2023specializing,han2024token,kang2024c3ot,xu2025chain,munkhbat2025selftraining}. Other efforts investigate compression through prompt compressors, gist tokens, latent-space reasoning, or broader test-time compute allocation \cite{jiang2023llmlingua,jiang2023longllmlingua,mu2023learning,hao2024training,snell2024scaling}. These approaches typically operate on reasoning content, supervision, decoding mechanisms, or hidden representations. Our work complements them by studying a linguistic register as the compression axis, without requiring architectural changes or specialized tokenizers.

\paragraph{Multilingual reasoning and language-transfer effects.}
A growing body of work studies how the language of reasoning interacts with reasoning quality in multilingual LLMs \cite{shi2022language}. Prior studies show that reasoning language, prompt language, cross-lingual transfer, and multilingual instruction tuning can affect both accuracy and consistency \cite{huang2023not,lai2024mcot,zhao2026comprehensive}. These studies mostly compare modern languages. Our question is different: whether a historically compact written variety of the same language family can serve as an efficient visible-reasoning register for models whose default reasoning language is modern Chinese.

\paragraph{Classical Chinese in NLP.}
Prior NLP work on Classical Chinese has largely focused on tasks such as machine translation between ancient and modern Chinese, Classical Chinese understanding benchmarks, and specialized ancient-text LLMs \cite{zhang2018automatic,zhou2023wyweb,wei2024aceval,cao2024tonggu,yao2025wenyangpt}. These efforts treat wenyan as the object of study rather than as a medium for model reasoning. Our work proposes a different role for Classical Chinese: as a compact, high-density surface channel for the visible reasoning of LLMs.

\paragraph{Chinese reasoning models.}
Recent Chinese-capable model families such as Qwen and DeepSeek have rapidly improved general, mathematical, and reasoning-oriented capabilities \cite{yang2024qwen25,shao2024deepseekmath,deepseek2024v3,guo2025deepseek,yang2025qwen3}. However, to our knowledge, prior work has not systematically asked whether a Chinese model's visible reasoning channel should default to modern vernacular Chinese, nor whether shifting it to wenyan can improve efficiency. Our work opens this conversation.

\section{Method}

\subsection{Task Formulation}

Let a question be denoted by $x$, a visible reasoning trace by $r$, and the final answer by $a$. In standard Chinese thinking models, the visible reasoning trace is typically generated in modern Chinese, followed by a short answer string. In contrast, our target format is
\[
x \mapsto (r_{\mathrm{wenyan}}, a_{\mathrm{direct}}),
\]
where the intermediate reasoning is expressed in Classical Chinese while the final answer remains in a short benchmark-required direct-answer format.

The reasoning trace is typically the most verbose part of the output and therefore the natural place to save tokens, while the final answer should remain short, evaluation-compatible, and amenable to exact-match evaluation. Our central hypothesis is $|r_{\mathrm{wenyan}}| < |r_{\mathrm{modern}}|$ while preserving answer quality; if this holds, total generation cost decreases even with an unchanged answer channel.

For analysis, we additionally construct a compact modern-Chinese control $x \mapsto (r_{\mathrm{compact}}, a_{\mathrm{direct}})$, where $r_{\mathrm{compact}}$ is a concise modern-Chinese rewrite of the same source reasoning trace. This control uses the same answer channel and training recipe, and disentangles the effect of generic shortening from the effect of shifting the reasoning register to wenyan.

\begin{figure*}[t]
    \centering
    \includegraphics[width=\textwidth]{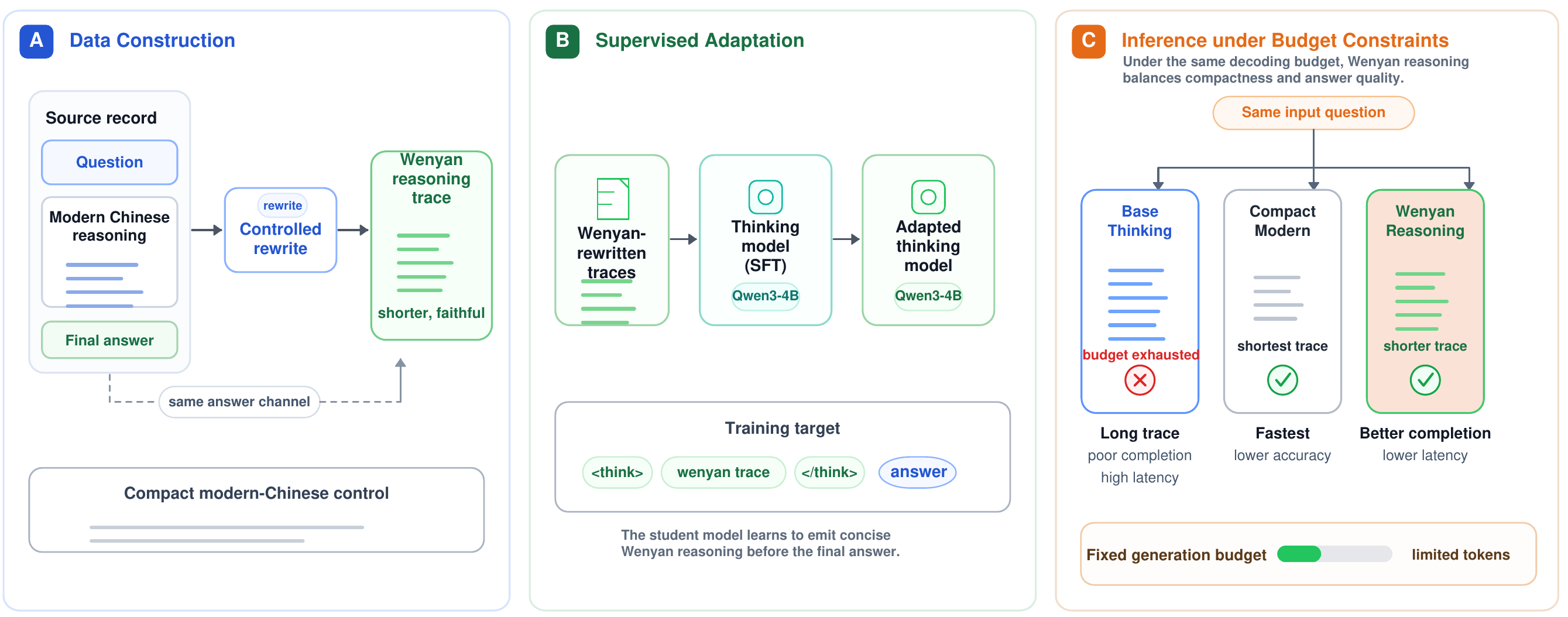}
    \caption{Overview of \method. We first convert modern Chinese reasoning traces into concise Wenyan-style rewrites through controlled rewriting, then perform supervised fine-tuning on the Wenyan-rewritten reasoning data to obtain a Wenyan-thinking model. Under limited generation budgets, the resulting model emits shorter reasoning traces, leaving more room for final answer completion and improving decoding efficiency.}
    \label{fig:wenyan_overview}
\end{figure*}

\subsection{Data Construction}

We begin with \textsc{Chinese-DeepSeek-R1-Distill-data-110k}, whose records contain a question, a final answer, and a modern-Chinese reasoning field \texttt{reasoning\_content}. We select 10,000 source examples for rewriting and require all retained examples to have a non-empty question, final answer, original reasoning trace, source-repository tag, teacher identifier, and rewrite status. We do not generate new answers. For the main wenyan setting, we rewrite only the reasoning field into Classical Chinese while leaving the question and final answer unchanged.

To build a controlled comparison, we additionally construct a compact modern-Chinese version from the same source examples, where the reasoning field is rewritten into concise modern written Chinese rather than wenyan. This control uses the same input question, final answer, teacher family, and downstream SFT format as the wenyan setting.

Our main rewrite setting uses Qwen3.5-397B-A17B with the default wenyan prompt, which rewrites modern-Chinese reasoning traces into a compact classical register while preserving the overall step structure. We use Qwen3.5-397B-A17B as the primary rewriting teacher because it is a strong Chinese instruction-following model and produces stable structured rewrites under our prompt format. Alternative teachers and prompt styles, including DeepSeek and MiniMax variants, are reported in Appendix~\ref{app:mmlu_full_leaderboard} as sensitivity analyses rather than as separate main methods.

The rewrite outputs are stored as JSONL records with a dedicated \texttt{thinking\_wenyan} field. Concurrent API generation initially marked 87 of the 10,000 records as API failures; these records were repaired or regenerated before training, leaving 10,000 usable records with all required fields populated. In the final training JSONL, the original modern-Chinese reasoning field has a mean length of 1457.8 Chinese characters (median 926), while the rewritten \texttt{thinking\_wenyan} field has a mean length of 420.9 characters (median 382), confirming that the supervision is substantially shorter before any student-model decoding effects. The source \texttt{answer} field often contains solution-style text rather than a benchmark option label; benchmark evaluations separately use direct-answer prompts and exact-match extractors.

\subsection{Supervised Fine-Tuning}

We fine-tune \think with full-parameter supervised fine-tuning. For each training example, the prompt is formed from the user question, and the completion is constructed so that the model emits a rewritten reasoning span followed by the preserved answer span. Only completion tokens are supervised. Benchmark-time prompts then require the short direct-answer format used by each evaluation task. This makes the training objective compatible with the original thinking-style interface of the base model while changing the linguistic register of the reasoning channel.

Our main implementation uses memory-optimized full-SFT with DeepSpeed ZeRO-3, bf16, gradient checkpointing, and a maximum sequence length of 2048. The final main system is trained for two epochs with AdamW-style optimization. Because the exact GPU topology can vary across servers, we report the optimization configuration rather than tie the method to a single hardware setup.

The compact modern-Chinese control is fine-tuned with the same recipe and answer format; the only difference is that the supervised reasoning span is a concise modern-Chinese rewrite rather than a wenyan rewrite. Thus, comparisons between \compact and \method control for source data, base model, answer channel, and SFT format.

Table~\ref{tab:data_train_setup} summarizes the resulting field lengths and the training recipe used for the final main system.

\begin{table}[t]
\centering
\small
\setlength{\tabcolsep}{4.5pt}
\begin{tabular}{lrrrr}
\toprule
Field & Mean & Median & Min & Max \\
\midrule
Question & 69.8 & 41 & 3 & 5811 \\
Answer & 905.3 & 799 & 4 & 7469 \\
Modern reasoning & 1457.8 & 926 & 41 & 24025 \\
Wenyan reasoning & 420.9 & 382 & 4 & 5794 \\
\bottomrule
\end{tabular}

\vspace{3pt}

\begin{tabular}{lc}
\toprule
Hyperparameter & Value \\
\midrule
Base model & \think \\
Training type & Full-parameter SFT \\
Max length & 2048 \\
Epochs & 2 \\
Learning rate & $1\times 10^{-5}$ \\
Weight decay & 0.1 \\
Warmup ratio & 0.03 \\
Optimizer & AdamW \\
Precision & bf16 \\
Gradient checkpointing & enabled \\
DeepSpeed & ZeRO-3 \\
\bottomrule
\end{tabular}
\caption{Training data statistics and main SFT configuration. The upper panel reports character lengths in \textsc{WenyanThinking-10k}; the lower panel reports the full-parameter SFT setup used by the final main system.}
\label{tab:data_train_setup}
\end{table}

\section{Experimental Setup}

\subsection{Benchmarks}

Our primary budget-sensitive evaluation is on \textsc{MMLU-ProX-Lite-ZH} in the answer-only setting. This is a fixed 588-example Chinese subset of MMLU-ProX \cite{xuan2025mmluprox}, which builds on the more difficult MMLU-Pro line of evaluation beyond the original MMLU benchmark \cite{hendrycks2021measuring,wang2024mmlupro}. The subset is designed for rapid budget sweeps over many model variants while retaining broad subject coverage across 14 subject groups. We call it ``Lite'' because it is a stratified evaluation subset rather than the full MMLU-ProX benchmark; all systems are evaluated on the same examples and with the same answer extractor. We further evaluate on \textsc{CEVAL-valid-direct} and an \textsc{AGIEval-CN} direct subset under the same generation budgets \cite{huang2023ceval,zhong2023agieval}; CMMLU provides a related large-scale reference point for Chinese multitask evaluation \cite{li2023cmmlu}. Appendix~\ref{app:data_eval_details} gives the subject counts, AGIEval task list, and decontamination audit.

\subsection{Compared Systems}

We distinguish \emph{main systems} from \emph{supporting variants}. The main systems are:
\begin{enumerate}[leftmargin=*,itemsep=1pt]
    \item \textbf{Think}: the original \think model.
    \item \textbf{Answer-only}: a Qwen3-4B-Instruct sibling evaluated with no visible thinking trace, testing whether direct answering alone explains the gains.
    \item \textbf{\concise}: the original \think model prompted to reason concisely without additional training, testing whether prompting alone can recover the budget-efficiency effect.
    \item \textbf{\vern}: SFT on long modern-Chinese reasoning traces.
    \item \textbf{\compact}: a controlled compression baseline trained on concise modern-Chinese rewrites from the same source data.
    \item \textbf{\method} (\ours): our main system, trained on wenyan-rewritten reasoning traces with the same final-answer format.
\end{enumerate}
In supporting analyses, we additionally compare teacher-derived wenyan variants (DeepSeek-V3.2, DeepSeek-V4-Flash, MiniMax-M2.7, Merged teachers) and Qwen3.5 prompt-style variants (faithful, ultra-compact) to study the sensitivity of the data construction stage.

\subsection{Budget-Sensitive Evaluation and Metrics}

We vary \texttt{max\_gen\_toks} over four representative budgets: 1024, 2048, 4096, and 6144, rather than treating it as a fixed background parameter. This is critical for reasoning models, since shorter traces can substantially improve answer completion under practical decoding limits. We use deterministic decoding (\texttt{do\_sample=False}, temperature 0) and report task accuracy together with generation-efficiency indicators: average emitted reasoning length, average total output length, missing-answer rate, and end-to-end evaluation time. The central efficiency question is whether models can think more economically under realistic limits, not merely whether they can think longer.

For GPU-based latency measurements, all compared systems in the same table are evaluated with the same lm-evaluation-harness and vLLM setup \cite{kwon2023efficient}: bf16 inference, tensor parallel size 1, \texttt{max\_model\_len}=8192, automatic batching, and identical deterministic decoding. We report total wall-clock evaluation time as a deployment-oriented throughput indicator rather than a hardware-independent per-token constant.

\section{Main Results on the Primary Benchmark}

\subsection{Overall Comparison Under Different Generation Budgets}

Table~\ref{tab:mmlu_main} reports exact-match accuracy on \textsc{MMLU-ProX-Lite-ZH} under different generation budgets. Unless otherwise specified, \textbf{Wenyan} refers to our main Qwen3.5-397B-A17B default wenyan rewrite setting.

\begin{table}[t]
\centering
\small
\setlength{\tabcolsep}{5pt}
\begin{tabular}{lcccc}
\toprule
Model & 1024 & 2048 & 4096 & 6144 \\
\midrule
Answer-only & 41.5 & 41.5 & 41.0 & 41.5 \\
Think & 11.4 & 32.3 & 50.9 & \textbf{58.8} \\
Concise prompt & 13.4 & 32.0 & 44.9 & 47.5 \\
Long modern SFT & 37.1 & 47.8 & 51.2 & 53.7 \\
Compact modern & \underline{52.2} & \underline{52.9} & \underline{53.2} & 53.4 \\
Wenyan & \textbf{52.9} & \textbf{55.3} & \textbf{55.4} & \underline{55.6} \\
\bottomrule
\end{tabular}
\caption{Exact-match (\%) on \textsc{MMLU-ProX-Lite-ZH} under different generation budgets. Wenyan denotes our final Qwen3.5-based main system. Concise prompt is a prompt-only baseline without SFT; Answer-only uses no visible thinking trace. \textbf{Boldface marks the best accuracy, and underline marks the second best for each budget column.}}
\label{tab:mmlu_main}
\end{table}

Table~\ref{tab:mmlu_main} reveals a clear budget-sensitive trade-off. \ours substantially improves over \think under short and medium budgets (11.4 $\to$ 52.9 at 1024; 32.3 $\to$ 55.3 at 2048; 50.9 $\to$ 55.4 at 4096). The practical advantage is that wenyan reasoning reaches strong accuracy much earlier in the generation budget, before long-form thinking has enough room to finish its trace and answer.

The controls clarify what drives the effect. Direct answering without a visible thinking trace is stable but capped around 41\%, so the gains do not come from simply suppressing reasoning. Prompting \think to be concise helps at larger budgets but remains far below trained compact-reasoning systems, showing that prompt-only shortening is insufficient. Finally, \compact is already strong, confirming that compact rewriting is an important part of the mechanism. \ours is consistently above this compact modern-Chinese control on the primary benchmark, but the margin is modest; we therefore interpret wenyan as a viable compact register that can trade slightly longer traces for higher accuracy, rather than as a universally superior alternative to compact modern Chinese.

\subsection{Subject-Level Behavior}

Table~\ref{tab:subjectwise_main} reports representative subject-wise results at the strictest 1024-token budget and the least constrained 6144-token budget. The 1024-token setting is most diagnostic for our method: the base thinking model often spends the entire budget on unfinished reasoning, while compact-reasoning systems can complete both reasoning and answer.

\begin{table*}[t]
\centering
\small
\setlength{\tabcolsep}{4.3pt}
\begin{tabular}{lcccccccc}
\toprule
& \multicolumn{4}{c}{Budget = 1024} & \multicolumn{4}{c}{Budget = 6144} \\
\cmidrule(lr){2-5}\cmidrule(lr){6-9}
Subject & Think & Long & Compact & Wenyan & Think & Long & Compact & Wenyan \\
\midrule
Business & 2.5 & 42.5 & 60.0 & \textbf{65.0} & \textbf{65.0} & 60.0 & 60.0 & \textbf{65.0} \\
Chemistry & 5.4 & 30.4 & \textbf{55.4} & 53.6 & 64.3 & \textbf{66.1} & 57.1 & 62.5 \\
Computer Sci. & 5.0 & 20.0 & 45.0 & \textbf{55.0} & \textbf{65.0} & 40.0 & 45.0 & 60.0 \\
Economics & 23.8 & 59.5 & 66.7 & \textbf{73.8} & \textbf{73.8} & \textbf{73.8} & 66.7 & \textbf{73.8} \\
Law & 0.0 & 2.1 & 18.8 & \textbf{31.2} & 27.1 & 6.2 & 20.8 & \textbf{31.2} \\
Math & 29.4 & 51.5 & \textbf{69.1} & \textbf{69.1} & \textbf{83.8} & 82.4 & 70.6 & 72.1 \\
Physics & 13.8 & 49.2 & 64.6 & \textbf{69.2} & 67.7 & 66.2 & 67.7 & \textbf{73.8} \\
Psychology & 15.0 & 47.5 & \textbf{65.0} & 52.5 & 65.0 & \textbf{67.5} & 65.0 & 52.5 \\
\midrule
\textbf{Overall} & 11.4 & 37.1 & 52.2 & \textbf{52.9} & \textbf{58.8} & 53.7 & 53.4 & 55.6 \\
\bottomrule
\end{tabular}
\caption{Representative subject-wise exact-match (\%) on \textsc{MMLU-ProX-Lite-ZH}. Long denotes long modern-Chinese SFT; Compact denotes the compact modern-Chinese control; Wenyan denotes our final Qwen3.5-based main system. \textbf{Boldface marks the best accuracy within each subject and budget block.}}
\label{tab:subjectwise_main}
\end{table*}

At 1024 tokens, \ours outperforms long-form thinking across all shown subjects and is especially strong on economics, law, physics, business, and computer science. At 6144 tokens, long-form thinking recovers on some subjects, especially math, but \ours remains competitive while using much shorter traces. This subject-level pattern supports the main interpretation: wenyan supervision is most valuable when reasoning budget is scarce.

\subsection{Efficiency and Answer Completion}

Table~\ref{tab:efficiency_main} analyzes emitted reasoning length and answer completion. The operational mechanism is clear: at 1024 tokens, \think emits 1008.1 output tokens with an 87.5\% missing-answer rate (11.4\% accuracy), while \ours uses 394.2 tokens with a 7.5\% missing rate (52.9\% accuracy). At 4096 tokens, \ours uses 458.4 tokens versus 2901.8 for the base model. The compact modern-Chinese control is even shorter and has lower missing-answer rates, while Wenyan obtains higher accuracy on the primary benchmark. This points to an accuracy--compactness trade-off rather than pure token minimization.

\begin{table*}[t]
\centering
\small
\setlength{\tabcolsep}{4.8pt}
\begin{tabular}{lccccccccc}
\toprule
& \multicolumn{3}{c}{Accuracy (\%)} & \multicolumn{3}{c}{Avg. output tokens} & \multicolumn{3}{c}{Missing-answer rate} \\
\cmidrule(lr){2-4}\cmidrule(lr){5-7}\cmidrule(lr){8-10}
Budget & Think & Compact & Wenyan & Think & Compact & Wenyan & Think & Compact & Wenyan \\
\midrule
1024 & 11.4 & 52.2 & \textbf{52.9} & 1008.1 & \textbf{254.7} & 394.2 & 87.5\% & \textbf{3.1\%} & 7.5\% \\
2048 & 32.3 & 52.9 & \textbf{55.3} & 1859.4 & \textbf{267.5} & 434.3 & 59.9\% & \textbf{0.7\%} & 1.9\% \\
4096 & 50.9 & 53.2 & \textbf{55.4} & 2901.8 & \textbf{274.0} & 458.4 & 27.1\% & \textbf{0.2\%} & 0.7\% \\
6144 & \textbf{58.8} & 53.4 & 55.6 & 3389.7 & \textbf{274.1} & 467.9 & 14.3\% & \textbf{0.0\%} & 0.3\% \\
\bottomrule
\end{tabular}
\caption{Length and answer-completion analysis under different generation budgets. Compact denotes the modern-Chinese compact control, and Wenyan denotes our final Qwen3.5-based main system. \textbf{Boldface marks the best value within each metric block: highest accuracy and lowest tokens/rates.}}
\label{tab:efficiency_main}
\end{table*}

\subsection{Does the Answer Channel Drive the Gain?}
\label{sec:ablation_train_target}

To verify that the method is not merely learning benchmark answer formatting, we compare the default full-completion SFT objective with a variant that applies loss only to the reasoning span up to \texttt{</think>}. Both use the same base model and the same wenyan-rewritten data.

The default objective reaches 52.9, 55.3, 55.4, 55.6 across the 1024, 2048, 4096, 6144 budgets, while the reasoning-only variant reaches 52.4, 54.1, 54.6, 54.9. Removing answer-token supervision therefore causes only small drops of 0.5-1.2 points. This indicates that the main benefit comes from learning a compact and stable wenyan reasoning style, while direct answer supervision modestly stabilizes the transition from reasoning to final answer.

\subsection{LLM-as-Judge Audit of Rewrite Quality}
\label{sec:llm_judge_quality}

Because the method depends on teacher-generated rewriting, we further audit whether wenyan rewrites preserve the original reasoning rather than merely exposing answer cues. We sample 40 rewritten training examples and ask DeepSeek-V4-Pro, used only as an automatic judge, to score each rewrite on faithfulness, step preservation, readability/reversibility, answer support, answer-leakage risk, and over-compression loss. The judge sees the question, final answer, original modern-Chinese reasoning, and the wenyan rewrite, and returns structured JSON scores. Appendix~\ref{app:llm_judge_quality} gives the audit protocol and additional details.

\begin{table}[t]
\centering
\small
\setlength{\tabcolsep}{3.8pt}
\resizebox{\linewidth}{!}{
\begin{tabular}{lrrrrrrr}
\toprule
$n$ & Faith. & Steps & Read. & Support & Leak. $\downarrow$ & Loss $\downarrow$ & Major err. \\
\midrule
40 & 4.35 & 4.30 & 4.33 & 4.45 & 1.00 & 1.53 & 0.00 \\
\bottomrule
\end{tabular}
}
\caption{LLM-as-judge audit of wenyan rewrite quality. Scores are on 1--5 scales; higher is better except leakage risk and over-compression loss.}
\label{tab:llm_judge_quality}
\end{table}

Table~\ref{tab:llm_judge_quality} suggests that the rewrites are usually faithful and auditable: the mean faithfulness, step-preservation, readability, and answer-support scores are all above 4.3, while answer-leakage risk is minimal (1.00) and the judge marks no major-error cases. Verdicts are 27 pass, 11 minor issue, and 2 major issue; the non-pass cases are mostly omissions in open-ended explanatory examples rather than answer-changing errors. We treat this as supporting evidence, not a replacement for human annotation.

\paragraph{Additional analyses.}
Appendix~\ref{app:mmlu_full_leaderboard} reports teacher-source and prompt-style sensitivity, Appendix~\ref{app:llm_judge_quality} reports the rewrite-quality audit, and Appendix~\ref{app:reasoning_case_study} provides an annotated reasoning-trace comparison. These results support the same interpretation: effective supervision requires compactness, step preservation, and stylistic consistency.

\section{Cross-Benchmark Validation}

\subsection{CEVAL}

To test whether the same budget-sensitive behavior generalizes beyond \textsc{MMLU-ProX-Lite-ZH}, we further evaluate on \textsc{CEVAL-valid-direct} under the same four generation budgets.

\begin{table}[t]
\centering
\small
\setlength{\tabcolsep}{4.8pt}
\begin{tabular}{lcccc}
\toprule
Model & 1024 & 2048 & 4096 & 6144 \\
\midrule
Think & 41.0 & 63.0 & \textbf{75.6} & \textbf{79.3} \\
Long modern & 63.4 & 69.4 & 70.6 & 73.0 \\
Compact modern & \underline{72.7} & \underline{72.8} & 72.9 & 73.1 \\
Wenyan & \textbf{73.5} & \textbf{74.8} & \underline{74.7} & \underline{74.7} \\
\bottomrule
\end{tabular}
\caption{Exact-match (\%) on \textsc{CEVAL-valid-direct} under different generation budgets. Compact modern rewrites the reasoning traces into concise modern Chinese rather than wenyan. \textbf{Boldface marks the best accuracy, and underline marks the second best for each budget column.}}
\label{tab:ceval_main}
\end{table}

CEVAL confirms the same budget-sensitive pattern (Table~\ref{tab:ceval_main}). Wenyan is strong at 1024--2048 tokens and remains competitive at larger budgets, while the original \think model needs 4096--6144 tokens to overtake compact-reasoning systems. Thus, the main benefit is early answer completion under realistic budgets.

\subsection{AGIEval-CN}

To test whether the observations from CEVAL transfer beyond a single benchmark, we further evaluate on the \textsc{AGIEval-CN} direct subset described in Section~4.1. Table~\ref{tab:agieval_summary} reports overall averages and end-to-end evaluation time under four generation budgets.

\begin{table}[t]
\centering
\small
\setlength{\tabcolsep}{4.5pt}
\resizebox{\columnwidth}{!}{
\begin{tabular}{lcccccccc}
\toprule
& \multicolumn{4}{c}{Accuracy (\%)} & \multicolumn{4}{c}{Time (Latency)} \\
\cmidrule(lr){2-5}\cmidrule(lr){6-9}
Model & 1024 & 2048 & 4096 & 6144 & 1024 & 2048 & 4096 & 6144 \\
\midrule
Think & 14.20 & 24.37 & 33.49 & 35.89 & 24:20 & 1:06:19 & 2:35:33 & 3:36:37 \\
Long modern & 40.89 & 47.82 & \textbf{52.22} & \textbf{52.97} & 19:59 & 40:59 & 1:39:32 & 3:02:33 \\
Compact modern & \underline{48.94} & \underline{49.45} & 49.80 & 49.83 & \textbf{05:28} & \textbf{06:11} & \textbf{06:48} & \textbf{07:34} \\
Wenyan & \textbf{49.92} & \textbf{51.55} & \underline{51.43} & \underline{51.27} & \underline{07:54} & \underline{08:53} & \underline{10:14} & \underline{11:20} \\
\bottomrule
\end{tabular}
}
\caption{Zero-shot accuracy (\%) and total evaluation time on the \textsc{AGIEval-CN} direct subset under different generation budgets. The subset excludes Gaokao MathCloze. \textbf{Boldface marks the best (highest accuracy, lowest latency), and underline marks the second best for each budget column.}}
\label{tab:agieval_summary}
\end{table}

AGIEval-CN further supports the efficiency story (Table~\ref{tab:agieval_summary}). Compact-reasoning systems dramatically reduce evaluation time while improving accuracy under tight budgets: Wenyan reaches 51.43 at 4096 tokens in 10:14 total evaluation time, compared with 33.49 and 2:35:33 for the base thinking model. Compact modern Chinese is faster, while Wenyan trades additional generation time for higher accuracy at 1024--2048 tokens. The appendix reports the full task-level table and additional teacher variants.

\section{Discussion}

\subsection{Main Takeaways and Interpretation}

Our results support three central findings. First, the reasoning and answer channels can be decoupled in format: a Chinese thinking model can reason in one linguistic register while answering in another, without architectural changes. Second, compact, logic-preserving rewriting shortens generated reasoning traces and improves the accuracy--budget--latency trade-off under constrained decoding. Third, the compact modern-Chinese control shows that shortening is an important mechanism, while the Qwen3.5-based wenyan system shows that a linguistically grounded register can serve as a strong compact reasoning channel. The Wenyan--Compact gap should be read as a modest accuracy advantage on our primary benchmark, not as proof that wenyan is always the shortest or universally best register.

The strongest implication is broader than any single teacher model: for Chinese LLMs with visible reasoning traces, the \emph{linguistic register} of the reasoning channel is a meaningful and previously overlooked design variable for reasoning efficiency. The supporting analyses further sharpen this interpretation: good supervision must be compact, learnable, and faithful to the original reasoning steps. They also show that teacher choice and rewrite style can dominate register effects on some benchmarks, so register shift should be treated as a design axis rather than as a one-size-fits-all recipe.

The comparison with \compact is especially informative. Compact modern Chinese is often the shortest and has the lowest missing-answer rate, but the Wenyan model obtains higher accuracy on the primary benchmark and in the low-budget CEVAL setting. We therefore view wenyan as a compact-but-expressive point on the compression frontier: it removes much of the discourse overhead of long modern-Chinese reasoning while still preserving enough connective structure for the student model to learn a stable reasoning channel. This is the paper's intended claim, not that classical Chinese is intrinsically optimal for every task or budget.

\subsection{Broader Implications and Risks}

Our findings suggest that register selection for reasoning could extend beyond Chinese: other languages with diglossic or historical varieties may offer analogous compact reasoning registers. More broadly, the surface form of intermediate computation in LMs deserves systematic study alongside the dominant focus on content and correctness.

\section{Conclusion}

This paper studies whether Chinese LLMs can ``think in wenyan.'' We propose \method, a training setting in which the model emits visible intermediate reasoning in Classical Chinese while keeping the final answer in a short direct-answer format. Across three Chinese reasoning benchmarks under controlled generation budgets, \method shortens emitted traces, reduces missing-answer rates, and yields a strong accuracy--budget--latency trade-off. A compact modern-Chinese control confirms that shortening is central; the wenyan system further shows that linguistic register is a practical design axis for efficient visible reasoning, offering a competitive compact register that often trades slightly longer traces for higher accuracy.

\section*{Limitations}

Our study focuses on Chinese reasoning benchmarks and a 10,000-example rewritten training subset; broader task coverage and additional base model families remain useful future work. The primary benchmark is a 588-example Lite subset intended for budget sweeps, so larger-scale evaluation would better estimate small gaps between compact registers. Because the supervision is teacher-generated, rewrite quality and consistency matter, and compact reasoning can be harder to audit than longer vernacular traces; our LLM-as-judge audit provides an automatic check, but not a substitute for full human evaluation of faithfulness, step preservation, or readability. The answer-only baseline uses an instruct sibling rather than a strictly identical no-thinking decoding of the same base model. Finally, the current evidence is strongest for Chinese, where wenyan provides a naturally compact historical register; whether analogous registers help in other languages is left open.

\bibliography{custom}

\appendix

\section{Additional Results}
\label{app:additional_results}

\subsection{Data, Evaluation, and Decontamination Details}
\label{app:data_eval_details}

\textsc{MMLU-ProX-Lite-ZH} contains 588 examples across 14 subject groups: biology (36), business (40), chemistry (56), computer science (20), economics (42), engineering (48), health (35), history (19), law (48), math (68), other (46), philosophy (25), physics (65), and psychology (40). The AGIEval-CN direct subset includes Gaokao Biology, Chemistry, Chinese, Geography, History, MathQA, Physics, JEC-QA-CA, JEC-QA-KD, and LogiQA-ZH; Gaokao MathCloze is excluded because its fill-in format does not match the unified direct-answer protocol.

For decontamination, we run exact-question matching and 8-gram lexical-overlap matching between the 10,000 rewritten training records and collected evaluation question files. Across 296,681 scanned evaluation questions, exact question overlap is zero. The high-overlap screen at Jaccard threshold 0.75 flags repeated logs of a single AGIEval-style geography example from \texttt{exam/coig\_exam}; it does not affect the primary MMLU-ProX-Lite-ZH evaluation.

This appendix provides complete result tables and supplementary figures that support the main paper.
Consistent with the main text, we separate the primary claim from supporting sensitivity analyses:
the main story concerns whether compact, logic-preserving Wenyan reasoning improves the accuracy--budget trade-off, whereas prompt-style and teacher-source comparisons are used to understand robustness and boundary conditions.

\subsection{LLM-as-Judge Rewrite Quality Audit}
\label{app:llm_judge_quality}

To complement automatic task accuracy, we run a small LLM-as-judge audit on 40 randomly sampled rewritten training examples. The judge is DeepSeek-V4-Pro, which is not used as the main rewriting teacher for our Qwen3.5-based system. For each example, the judge receives the question, final answer, original modern-Chinese reasoning trace, and the wenyan rewrite. It scores six dimensions on 1--5 scales: faithfulness, step preservation, readability/reversibility, answer support, answer-leakage risk, and over-compression loss. Higher is better for the first four dimensions; lower is better for leakage and over-compression. We additionally ask for a binary major-error judgment and a pass/minor/major verdict.

\begin{table}[htbp]
\centering
\small
\setlength{\tabcolsep}{4.2pt}
\resizebox{\linewidth}{!}{
\begin{tabular}{lrrrrrrrr}
\toprule
Variant & $n$ & Faith. & Steps & Read. & Support & Leak. $\downarrow$ & Loss $\downarrow$ & Major err. \\
\midrule
Wenyan rewrite & 40 & 4.35 & 4.30 & 4.33 & 4.45 & 1.00 & 1.53 & 0.00 \\
\bottomrule
\end{tabular}
}
\caption{LLM-as-judge rewrite-quality audit. Faith., Steps, Read., and Support denote faithfulness, step preservation, readability/reversibility, and whether the rewrite supports the final answer.}
\label{tab:llm_judge_quality_appendix}
\end{table}

The audit yields 27 pass, 11 minor-issue, and 2 major-issue verdicts, with no example marked as having a major factual or logical error. The two major-issue verdicts are cases where open-ended, exploratory original reasoning was compressed into a much more direct rewritten trace; these indicate that compact rewriting can lose some discourse-level search behavior even when it does not change the final answer. This motivates our conservative framing: wenyan rewriting is generally faithful in this audit, but it should be viewed as a compact reasoning register that trades verbosity for a denser and more auditable trace, not as a guarantee that every teacher rewrite preserves every deliberative detail.

\subsection{Complete Leaderboard of MMLU-ProX-Lite-ZH Over Evaluated Systems}
\label{app:mmlu_full_leaderboard}

Table~\ref{tab:full_leaderboard_all_models} reports the complete MMLU-ProX-Lite-ZH leaderboard over all evaluated systems discussed in the paper.
Unlike the grouped summary tables in the main text, this appendix table is intended as a compact reference covering the main system, compact modern-Chinese controls, prompt-style variants, and teacher-source variants.

\begin{table*}[htbp]
\centering
\scriptsize
\setlength{\tabcolsep}{4.2pt}
\begin{tabular}{llcccc}
\toprule
\multirow{2}{*}{Group} & \multirow{2}{*}{Model} & \multicolumn{4}{c}{Exact-match accuracy (\%)} \\
\cmidrule(lr){3-6}
& & 1024 & 2048 & 4096 & 6144 \\
\midrule
\multirow{5}{*}{Baselines / non-Wenyan}
& Qwen3-4B-Instruct-2507 answer-only & 41.5 & 41.5 & 41.0 & 41.5 \\
& Qwen3-4B-Thinking-2507 & 11.4 & 32.3 & 50.9 & \textbf{58.8} \\
& Qwen3-4B-Thinking-2507 concise prompt & 13.4 & 32.0 & 44.9 & 47.5 \\
& Baihua-SFT & 37.1 & 47.8 & 51.2 & 53.7 \\
& Compact modern Chinese control & 52.2 & 52.9 & 53.2 & 53.4 \\
\midrule
\multirow{7}{*}{Wenyan variants}
& DeepSeek-V3.2 normal & \textbf{53.7} & 53.7 & 54.1 & 55.1 \\
& \textbf{Ours: Qwen3.5-397B-A17B normal} & 52.9 & \textbf{55.3} & \textbf{55.4} & 55.6 \\
& Qwen3.5-397B-A17B faithful & 47.4 & 53.1 & 54.3 & 53.4 \\
& Qwen3.5-397B-A17B ultra compact & 49.3 & 49.5 & 51.2 & 51.0 \\
& MiniMax-M2.7 normal & 49.1 & 51.0 & 51.0 & 51.0 \\
& Merged teachers (DeepSeek-V3.2 / Qwen3.5-397B-A17B / MiniMax-M2.7) & 53.2 & 53.4 & 51.0 & 53.2 \\
& DeepSeek-V4-Flash normal & 51.7 & 52.4 & 54.1 & 54.1 \\
\bottomrule
\end{tabular}
\caption{Complete MMLU-ProX-Lite-ZH leaderboard across the evaluated systems. Values are exact-match accuracy (\%). The concise-prompt baseline is prompt-only and uses no additional SFT. The compact modern-Chinese control uses the same source examples, teacher family, base model, and SFT recipe as the Wenyan system, but rewrites the reasoning traces into concise modern Chinese rather than Classical Chinese. \textbf{Boldface marks the best accuracy.}}
\label{tab:full_leaderboard_all_models}
\end{table*}

\begin{table*}[htbp]
\centering
\scriptsize
\setlength{\tabcolsep}{4.2pt}
\begin{tabular}{llcccc}
\toprule
\multirow{2}{*}{Group} & \multirow{2}{*}{Model} & \multicolumn{4}{c}{Exact-match accuracy (\%)} \\
\cmidrule(lr){3-6}
& & 1024 & 2048 & 4096 & 6144 \\
\midrule
\multirow{3}{*}{Baselines / non-Wenyan}
& Qwen3-4B-Thinking-2507 & 41.0 & 63.0 & \textbf{75.6} & \textbf{79.3} \\
& Baihua-SFT & 63.4 & 69.4 & 70.6 & 73.0 \\
& Compact modern Chinese control & 72.7 & 72.8 & 72.9 & 73.1 \\
\midrule
\multirow{7}{*}{Wenyan variants}
& DeepSeek-V3.2 normal & 73.9 & 74.3 & 74.9 & 73.3 \\
& \textbf{Ours: Qwen3.5-397B-A17B normal} & 73.5 & 74.8 & 74.7 & 74.7 \\
& Qwen3.5-397B-A17B faithful & 71.0 & 72.7 & 72.7 & 72.8 \\
& Qwen3.5-397B-A17B ultra compact & 73.1 & 73.4 & 73.5 & 73.5 \\
& MiniMax-M2.7 normal & 73.8 & \textbf{74.9} & 74.1 & 74.9 \\
& Merged teachers (DeepSeek-V3.2 / Qwen3.5-397B-A17B / MiniMax-M2.7) & 72.6 & 72.7 & 72.2 & 72.4 \\
& DeepSeek-V4-Flash normal & \textbf{74.1} & 74.8 & 74.9 & 74.9 \\
\bottomrule
\end{tabular}
\caption{Complete CEVAL-valid-direct leaderboard across the evaluated systems used in the final paper. Values are zero-shot exact-match accuracy (\%). The compact modern-Chinese control uses concise modern Chinese reasoning traces rather than Wenyan, and is included to disentangle the effect of compact rewriting from the effect of shifting the reasoning register. \textbf{Boldface marks the best accuracy.}}
\label{tab:ceval_full_leaderboard_all_models}
\end{table*}

\subsection{Complete AGIEval-CN Results}
\label{app:agieval_full_results}

To complement the compact AGIEval-CN discussion in the main text, we report the full per-subtask table and the corresponding overall accuracy--latency summary here.
We use the AGIEval-CN direct subset, excluding Gaokao MathCloze because it does not fit the unified direct-answer evaluation protocol used in this paper.

\begin{table*}[t]
\centering
\scriptsize
\setlength{\tabcolsep}{2.8pt}
\resizebox{\textwidth}{!}{
\begin{tabular}{lcccccccccccccccccccc}
\toprule
\multirow{3}{*}{Category}
& \multicolumn{12}{c}{Non-Wenyan / Controls}
& \multicolumn{8}{c}{Wenyan Variants} \\
\cmidrule(lr){2-13}\cmidrule(lr){14-21}
& \multicolumn{4}{c}{Qwen3-4B-Thinking-2507}
& \multicolumn{4}{c}{Long Vernacular SFT}
& \multicolumn{4}{c}{Compact modern Chinese control}
& \multicolumn{4}{c}{Wenyan (DeepSeek-V4-Flash teacher)}
& \multicolumn{4}{c}{Wenyan Default (Ours)} \\
\cmidrule(lr){2-5}\cmidrule(lr){6-9}\cmidrule(lr){10-13}\cmidrule(lr){14-17}\cmidrule(lr){18-21}
& 1024 & 2048 & 4096 & 6144
& 1024 & 2048 & 4096 & 6144
& 1024 & 2048 & 4096 & 6144
& 1024 & 2048 & 4096 & 6144
& 1024 & 2048 & 4096 & 6144 \\
\midrule
\textbf{Avg.}
& 14.20 & 24.37 & 33.49 & 35.89
& 40.89 & 47.82 & \textbf{52.22} & \textbf{52.97}
& 48.94 & 49.45 & 49.80 & 49.83
& \textbf{52.15} & \textbf{52.69} & 52.08 & 51.87
& 49.92 & 51.55 & 51.43 & 51.27 \\
Gaokao Biology
& 17.14 & 24.29 & 30.00 & 30.00
& 74.29 & 77.62 & \textbf{79.52} & \textbf{79.52}
& 73.33 & 74.29 & 73.33 & 73.33
& \textbf{81.43} & \textbf{78.57} & 77.14 & 79.05
& 74.76 & \textbf{78.57} & 73.81 & 74.76 \\
Gaokao Chemistry
& 7.73 & 15.94 & 25.12 & 34.30
& 45.89 & 53.14 & \textbf{60.39} & \textbf{59.90}
& 53.14 & 51.69 & 52.17 & 52.17
& \textbf{57.00} & 57.49 & 56.52 & 54.11
& 54.11 & \textbf{59.42} & 56.04 & 57.49 \\
Gaokao Chinese
& 12.20 & 22.76 & 22.36 & 24.39
& 51.22 & 58.54 & \textbf{61.79} & \textbf{60.98}
& 55.69 & 55.28 & 54.88 & 54.88
& \textbf{63.01} & \textbf{61.79} & 57.32 & 60.16
& 54.47 & 58.54 & 55.28 & 54.47 \\
Gaokao Geography
& 33.17 & 59.80 & 71.86 & \textbf{73.87}
& 58.29 & 64.82 & \textbf{72.36} & 69.35
& \textbf{71.36} & 70.85 & 69.35 & 69.35
& 69.85 & \textbf{72.86} & \textbf{72.36} & 67.84
& 67.84 & 71.36 & 70.85 & 67.84 \\
Gaokao History
& 38.72 & 61.70 & 76.60 & 77.02
& 71.91 & 74.04 & 74.89 & 74.89
& 76.60 & \textbf{77.02} & 77.45 & \textbf{77.45}
& \textbf{77.45} & 75.74 & \textbf{78.30} & 73.19
& 74.47 & 72.34 & 73.62 & 71.91 \\
Gaokao MathQA
& 37.04 & 61.82 & \textbf{79.77} & \textbf{85.19}
& 37.61 & 60.68 & 77.78 & 80.91
& 63.82 & 68.95 & 70.37 & 70.66
& 70.37 & 72.08 & 75.78 & 75.50
& \textbf{71.79} & \textbf{76.92} & 79.20 & 76.64 \\
Gaokao Physics
& 21.50 & 42.00 & 68.00 & \textbf{73.50}
& 45.50 & 63.00 & \textbf{71.50} & 71.00
& 59.50 & 62.00 & 64.50 & 64.50
& 63.50 & \textbf{65.50} & 64.50 & 62.50
& \textbf{68.00} & \textbf{65.50} & 69.50 & 69.00 \\
JEC-QA-CA
& 5.61 & 10.11 & 19.72 & 21.52
& 22.32 & 25.83 & 28.23 & 28.63
& 28.23 & 29.03 & 28.93 & 28.93
& 30.33 & \textbf{30.83} & 30.03 & 29.53
& \textbf{30.83} & 30.53 & \textbf{30.53} & \textbf{31.63} \\
JEC-QA-KD
& 6.51 & 12.31 & 14.51 & 15.52
& 33.23 & 34.43 & 34.23 & 35.14
& 35.34 & 35.24 & 36.04 & 36.04
& \textbf{37.84} & \textbf{38.84} & \textbf{37.54} & \textbf{37.54}
& 35.94 & 36.04 & 36.54 & 35.94 \\
LogiQA-ZH
& 11.83 & 18.13 & 28.88 & 31.34
& 48.69 & 60.52 & \textbf{67.59} & \textbf{70.35}
& 61.75 & 60.83 & 61.14 & 61.14
& \textbf{64.67} & \textbf{65.28} & 64.52 & 66.97
& 57.91 & 63.59 & 61.75 & 62.52 \\
\bottomrule
\end{tabular}
}
\caption{Per-subtask zero-shot accuracy (\%) on the AGIEval-CN direct subset under different generation budgets. The subset excludes Gaokao MathCloze because it does not fit our unified direct-answer evaluation protocol. We separate non-Wenyan baselines/controls from Wenyan-based variants for clarity. \textbf{Boldface marks the best accuracy.}}
\label{tab:agieval_full_detailed}
\end{table*}

\begin{table*}[t]
\centering
\resizebox{\textwidth}{!}{%
\begin{tabular}{llcccccccc}
\toprule
\multirow{2}{*}{Group} & \multirow{2}{*}{Model} & \multicolumn{4}{c}{Accuracy (\%)} & \multicolumn{4}{c}{Time (Latency)} \\
\cmidrule(lr){3-6}\cmidrule(lr){7-10}
& & 1024 & 2048 & 4096 & 6144 & 1024 & 2048 & 4096 & 6144 \\
\midrule
\multirow{3}{*}{Non-Wenyan / Controls}
& Qwen3-4B-Thinking-2507 & 14.20 & 24.37 & 33.49 & 35.89 & 24:20 & 1:06:19 & 2:35:33 & 3:36:37 \\
& Long Vernacular SFT Reasoning & 40.89 & 47.82 & \textbf{52.22} & \textbf{52.97} & 19:59 & 40:59 & 1:39:32 & 3:02:33 \\
& Compact modern Chinese control & 48.94 & 49.45 & 49.80 & 49.83 & \textbf{05:28} & \textbf{06:11} & \textbf{06:48} & \textbf{07:34} \\
\midrule
\multirow{2}{*}{Wenyan Variants}
& Wenyan (DeepSeek-V4-Flash teacher) & \textbf{52.15} & \textbf{52.69} & 52.08 & 51.87 & 07:09 & 07:56 & 10:42 & 13:21 \\
& Wenyan Default (Ours) & 49.92 & 51.55 & 51.43 & 51.27 & 07:54 & 08:53 & 10:14 & 11:20 \\
\bottomrule
\end{tabular}%
}
\caption{Zero-shot accuracy (\%) and total evaluation time on the AGIEval-CN direct subset under different generation budgets. The subset excludes Gaokao MathCloze. We separate non-Wenyan baselines/controls from Wenyan-based variants for clarity. \textbf{Boldface marks the best (highest accuracy, lowest latency).}}
\label{tab:agieval_full_acc_time}
\end{table*}

\subsection{MMLU-ProX-Lite-ZH Full Subject-by-Budget Results of the Main System}
\label{app:mmlu_main_full_subjects}

To provide a complete view beyond the representative subjects discussed in the main text, Table~\ref{tab:mmlu_full_subject_budget_main} reports the full subject-wise results of the final main system, namely the Qwen3.5-397B-A17B normal Wenyan prompt setting, under all four generation budgets.
This table is intended as a consistency check for the main paper: it shows that the reported improvements are not limited to a hand-picked subset of subjects.

\begin{table}[htbp]
\centering
\small
\setlength{\tabcolsep}{5pt}
\begin{tabular}{lcccc}
\toprule
Subject & 1024 & 2048 & 4096 & 6144 \\
\midrule
Biology & \textbf{63.9} & \textbf{63.9} & \textbf{63.9} & \textbf{63.9} \\
Business & \textbf{65.0} & \textbf{65.0} & \textbf{65.0} & \textbf{65.0} \\
Chemistry & 53.6 & \textbf{62.5} & \textbf{62.5} & \textbf{62.5} \\
Computer Science & 55.0 & \textbf{60.0} & \textbf{60.0} & \textbf{60.0} \\
Economics & \textbf{73.8} & \textbf{73.8} & \textbf{73.8} & \textbf{73.8} \\
Engineering & 25.0 & 29.2 & 29.2 & \textbf{31.2} \\
Health & \textbf{48.6} & \textbf{48.6} & \textbf{48.6} & \textbf{48.6} \\
History & \textbf{36.8} & \textbf{36.8} & \textbf{36.8} & \textbf{36.8} \\
Law & \textbf{31.2} & \textbf{31.2} & \textbf{31.2} & \textbf{31.2} \\
Math & 69.1 & 70.6 & \textbf{72.1} & \textbf{72.1} \\
Other & 32.6 & \textbf{37.0} & \textbf{37.0} & \textbf{37.0} \\
Philosophy & \textbf{44.0} & \textbf{44.0} & \textbf{44.0} & \textbf{44.0} \\
Physics & 69.2 & \textbf{73.8} & \textbf{73.8} & \textbf{73.8} \\
Psychology & \textbf{52.5} & \textbf{52.5} & \textbf{52.5} & \textbf{52.5} \\
\midrule
Overall & 52.9 & 55.3 & 55.4 & \textbf{55.6} \\
\bottomrule
\end{tabular}
\caption{Subject-wise exact-match (\%) of the final main system (Qwen3.5-397B-A17B normal Wenyan prompt) under all generation budgets. \textbf{Boldface marks the peak accuracy achieved for each subject across the different generation budgets.}}
\label{tab:mmlu_full_subject_budget_main}
\end{table}

\section{Prompt Templates}
\label{app:prompt_templates}

In this appendix, we present the three prompt templates used in our data generation (the wenyan thinking) pipeline.

\begin{tcolorbox}[promptzh, title={Prompt A (Chinese original) normal
controlled wenyan}]
\begin{Verbatim}[
  fontsize=\scriptsize,
  breaklines=true,
  breakanywhere=true,
  baselinestretch=0.95
]
system: |
  你现在要执行“受控文言化推理改写”任务。
  你的职责是：给定现代中文问题、现代中文推理过程和现代中文最终答案，只将“推理过程”改写为“适合推理模型训练的文言化推理”。
  
  你必须严格遵守以下要求：
  
  1. 只改写“现代推理”，不得输出问题，不得输出最终答案，不得输出任何额外说明。
  2. 输出必须是合法 JSON，且只包含一个字段：thinking_wenyan。
  3. 专业术语、英文缩写、模型名、数据集名、数字、单位、公式、变量名、代码、人名、地名必须保持原样。
  4. 不得将阿拉伯数字改写为中文数字，不得改写公式、符号与变量形式。
  5. 不得添加原文没有的新事实、新解释或新结论。
  6. 若原文存在多步推理，必须保留其逻辑顺序，不得改写为摘要式结论。
  7. 必须保留推出结论所必需的关键步骤、关键判断、关键比较、关键中间量与必要计算。
  8. 可以删除无信息增益的口语填充、重复确认、情绪语气、无贡献复查与重复表述。
  9. 可以合并意思完全重复的句子，但不得删除关键推理步骤。
  10. 文言化重点在逻辑连接、判断句、因果句、条件句、转折句与收束句；若某句完整文言化会降低可回译性，则保留现代书面表达，仅局部文言化。
  11. 不要使用生僻典故，不要故作艰深，不要把现代科技术语强行翻成古词。
  12. 输出风格以“稳、准、简、可回译”为先，应像紧凑的推理过程，而不是摘要、提纲或文艺化改写。
  
  额外约束：
  
  * 不保留“我想一想”“再看看”“似乎”“也许”“应当是”之类低信息犹疑语，除非该犹疑本身承载关键分支判断。
  * 不保留纯重复的验算与复述；但若复查产生新的判断信息，则应保留。
  * 不得把第一人称思考整体改写成客观说明文；应保持推理链条的过程感，但避免口语噪声。
  * 每一步尽量对应一个明确判断或一个明确操作，不要堆砌华丽文辞。

user: |
  问题：
  {question}
  
  现代推理：
  {reasoning}
  
  最终答案（仅供约束参考，不要输出）：
  {answer}
\end{Verbatim}
\end{tcolorbox}

\captionof{figure}{{Prompt A (Chinese original) normal controlled wenyan}
The original Chinese YAML prompt for the normal controlled wenyan
reasoning rewriting setting.}
\label{fig:prompt_normal_zh}

\begin{tcolorbox}[promptzh, title={Prompt B (Chinese original) faithful Wenyan}]
\begin{Verbatim}[
  fontsize=\scriptsize,
  breaklines=true,
  breakanywhere=true,
  baselinestretch=0.95
]
system: |
  你现在要执行“受控文言化推理改写（faithful）”任务。
  你的职责是：给定现代中文问题、现代中文推理过程和现代中文最终答案，只将“推理过程”改写为“适合推理模型训练的文言化推理”。

  本次改写的目标不是强压缩，也不是重写推理结构，而是：
  在尽量保持原推理步骤、原句子顺序、原信息粒度与原表达意图的前提下，
  仅做轻度文言化与必要去噪，使输出更稳定、更紧凑、但仍忠实于原始推理过程。

  你必须严格遵守以下要求：

  【一、内容边界】
  1. 只改写“现代推理”，不得输出问题，不得输出最终答案，不得输出任何额外说明。
  2. 输出必须是合法 JSON，且只包含一个字段：thinking_wenyan。
  3. 专业术语、英文缩写、模型名、数据集名、数字、单位、公式、变量名、代码、人名、地名必须保持原样。
  4. 不得将阿拉伯数字改写为中文数字，不得改写公式、符号与变量形式。
  5. 不得添加原文没有的新事实、新解释、新知识或新结论。

  【二、faithful 保留原则】
  6. 应尽量保留原推理的步骤顺序、句子边界、信息单位与表达层次，不得主动重组为更短的摘要式推理。
  7. 若原文存在多步推理，应尽量逐步对应保留，不得因追求简洁而删除原文中正常存在的推理展开。
  8. 必须保留关键步骤、关键判断、关键比较、关键中间量与必要计算。
  9. 即使某一步不是“最小充分推理链”中绝对不可少的一步，只要它属于原文自然的推理展开、且有助于保持过程感与可学习性，也应优先保留。
  10. 不主动压缩为“最小充分推理链”，不主动追求极限短写。
  11. 不主动合并相邻句子；只有当两句几乎完全重复、且合并后不改变原推理节奏时，才可合并。
  12. 不主动删除解释性短句；若其属于原推理自然组成部分，且并非明显冗余，应保留。
  13. 仅可删除明显无信息增益的内容，如：口语填充、纯情绪语气、无内容犹疑语、完全重复且不增加信息的复述、无贡献复查。
  14. 若某处复查产生新的判断信息、关键确认或关键排除，则应保留，不得删除。

  【三、文言化要求】
  15. 文言化重点在逻辑连接、判断句、因果句、条件句、转折句与收束句，可做局部替换，不必句句强行古化。
  16. 若某句完整文言化会降低可回译性，则保留现代书面表达，仅局部文言化。
  17. 不要使用生僻典故，不要故作艰深，不要把现代科技术语强行翻成古词。
  18. 输出风格以“稳、准、清、可回译”为先，应像原推理的忠实文言化版本，而非摘要、提纲、说明文或文学化改写。
  19. 不得把第一人称思考整体改写成客观说明文；应保持推理链条的过程感，但避免口语噪声。
  20. 每一步尽量对应一个明确判断或一个明确操作，不要堆砌华丽文辞。

  【四、最终判据】
  21. 改写后应与原推理在步骤数量、信息粒度和逻辑展开上大体对应，只做轻度文言化与轻度去噪。
  22. 改写后文本通常只应比原推理略短，不应显著缩短。
  23. 若原推理本身已经简洁，则主要做风格转换，不强行压缩。
  24. 优先保证“忠实保留原推理过程”，其次才是“语言更凝练”。

user: |
  问题：
  {question}

  现代推理：
  {reasoning}

  最终答案（仅供约束参考，不要输出）：
  {answer}
\end{Verbatim}
\end{tcolorbox}

\captionof{figure}{\textbf{Prompt B (Chinese original).}
The original Chinese YAML prompt for the faithful controlled wenyan
reasoning rewriting setting.}
\label{fig:prompt_faithful_zh}

\begin{tcolorbox}[promptzh, title={Prompt C (Chinese original) ultra-compact Wenyan}]
\begin{Verbatim}[
  fontsize=\scriptsize,
  breaklines=true,
  breakanywhere=true,
  baselinestretch=0.95
]
system: |
  你现在要执行“受控文言化推理改写（ultra_compact / 先秦简约风）”任务。
  你的职责是：给定现代中文问题、现代中文推理过程和现代中文最终答案，只将“推理过程”改写为“适合推理模型训练的高信息密度文言化最小充分推理链”。

  你的目标不是逐句翻译，也不是写文学化古文，而是：
  在不改变原推理结论依据的前提下，尽可能压缩表达，仅保留推出最终答案所必需的关键前提、关键判断、关键比较、关键中间量、必要计算与决定答案的分支选择，并以“先秦简约风”的方式表达：句短、义实、少饰、重判断、重因果、重转折、重收束。

  此处所谓“先秦简约风”，是指：
  借鉴先秦文辞之简约、凝练、少铺陈、重骨架的表达倾向；
  不是要求复原历史语言，不要求模仿具体典籍，不要求使用生僻古字或冷僻典故。

  你必须严格遵守以下要求：

  【一、内容边界】
  1. 只改写“现代推理”，不得输出问题，不得输出最终答案，不得输出任何额外说明。
  2. 输出必须是合法 JSON，且只包含一个字段：thinking_wenyan。
  3. 专业术语、英文缩写、模型名、数据集名、数字、单位、公式、变量名、代码、人名、地名必须保持原样。
  4. 不得将阿拉伯数字改写为中文数字，不得改写公式、符号与变量形式。
  5. 不得添加原文没有的新事实、新解释、新知识或新结论。
  6. 输出必须仍体现“推理过程”，不得直接压缩成只有一句结论。

  【二、ultra_compact 压缩原则】
  7. 只保留“推出答案不可缺少”的最小充分推理链；凡不影响结论成立者，皆删。
  8. 必须保留：关键条件、关键判断、关键比较、关键中间量、必要计算、关键分支与排除依据。
  9. 可以删除：口语填充、自问自答、重复确认、情绪语气、铺垫性解释、常识性展开、背景补充、无信息增益复查、纯复述性验算、仅增强流畅性的连接句。
  10. 若多句共同表达同一判断，可并为一句；若某一中间句不影响后续推出，可直接省去。
  11. 若存在多个分支，只保留最终实际采用且对答案必要的分支；未采用分支仅在“排除此分支”本身对答案必要时才可极简保留。
  12. 若某一步只是将上一步换言复述，应删除，不得保留同义重复。
  13. 若某处复查未产生新判断信息，应删除；若复查带来关键纠错或关键确认，则仅保留其最简形式。
  14. 优先保留“决定答案”的步骤，而非“解释给人看”的步骤。
  15. 若某一步仅用于增强自然语言连贯性，而非承担推理功能，应删除。

  【三、先秦简约风表达要求】
  16. 句式宜短，不拖沓；能断则断，能并则并。
  17. 多用判断、因果、条件、转折、递进、取舍等逻辑骨架；少作铺陈解释。
  18. 可适度使用“故、则、若、既、今、且、然、惟、是以、非…乃…、既…则…”等简洁连接，但不得刻意堆砌虚词。
  19. 不要使用生僻典故、冷僻异体字、艰深骈偶，不要故作古奥。
  20. 不要把现代科技术语强行翻成古词；凡术语、专名、公式、变量，保持原样。
  21. 若某句完整文言化会损害可回译性，则保留现代书面表达，仅局部文言化。
  22. 输出风格应“短、硬、稳、准”，像压缩后的推理骨架，而非文学化古文、摘要提纲或说明文。
  23. 每一句都应承载明确推理功能：给条件、作判断、作比较、算中间量、定取舍、收结论；不得写空泛过渡句。
  24. 可以适度省略不影响理解的主语、重复宾语或显然承接成分，但不得省略导致逻辑断裂。

  【四、最终压缩判据】
  25. 改写后文本应满足：若再删除其中任一步，原答案的推出依据就可能不完整。
  26. 改写后不得因过度压缩而丢失关键中间量、关键约束或关键分支判断。
  27. 宁少修饰，不漏关键一步。
  28. 最终输出应明显短于原推理，但仍可据此回译出核心推理链。
  29. 若原推理本已极短，只做必要压缩与文言化，不强求再缩至不自然。
  30. 若“更古、更短”的表达会损害训练可学性，则优先保证清楚、稳定、可回译，而非一味求古。

user: |
  问题：
  {question}

  现代推理：
  {reasoning}

  最终答案（仅供约束参考，不要输出）：
  {answer}
\end{Verbatim}
\end{tcolorbox}

\captionof{figure}{\textbf{Prompt C (Chinese original).}
The original Chinese YAML prompt for the ultra-compact controlled wenyan reasoning rewriting setting.}
\label{fig:prompt_ultracompact_zh}

\begin{tcolorbox}[promptzh, title={Prompt D (Chinese original): Compact Modern Chinese Control}]
\begin{Verbatim}[
  fontsize=\scriptsize,
  breaklines=true,
  breakanywhere=true,
  baselinestretch=0.95
]
system: |
  你现在要执行“受控紧凑现代中文推理改写”任务。
  你的职责是：给定现代中文问题、现代中文推理过程和现代中文最终答案，只将“推理过程”改写为“紧凑现代中文推理”。

  你必须严格遵守以下要求：
  1. 只改写现代推理，不改最终答案的语义。
  2. 专业术语、英文缩写、模型名、数据集名、数字、单位、公式、代码、人名地名全部保持原样。
  3. 不要使用文言文，不要使用古汉语表达，不要刻意文学化。
  4. 主要删除口语填充、重复确认、自问自答、无贡献复查、重复解释和冗余表述。
  5. 尽量压缩字数，但不得遗漏原推理中的关键步骤。
  6. 必须保留推出结论所需的关键条件、关键判断、关键比较、关键中间量和必要计算。
  7. 不得添加原文没有的新事实、新解释或新结论。
  8. 若原文中存在多步推理，必须保留其逻辑顺序。
  9. 若某些句子只是重复表达同一含义，可以合并为一句。
  10. 若某些内容只是口语化思考过程而不影响推理，可以删除。
  11. 若无法在不遗漏关键步骤的前提下进一步压缩，应优先保持原有现代书面表达稳定不变。
  12. 输出必须是合法 JSON，且只包含一个字段：thinking_wenyan。
  13. 注意：字段名 thinking_wenyan 只是为了兼容数据处理脚本；字段内容必须是紧凑现代中文推理，不得使用文言文。
  14. 不要输出最终答案，不要重复问题，不要输出任何额外说明，不要使用 markdown 代码块。

  风格要求：
  - 使用简洁、自然、规范的现代中文；
  - 以“稳”“准”“清楚”为先；
  - 允许保留必要解释，但删除明显冗余；
  - 输出应像压缩后的推理过程，而不是摘要或结论；
  - 不追求极限压缩，只需明显短于原推理。

  输出格式必须严格如下：
  {"thinking_wenyan":"这里填写改写后的紧凑现代中文推理"}

user: |
  问题：
  {question}

  现代推理：
  {reasoning}

  最终答案（仅供约束参考，不要输出）：
  {answer}
\end{Verbatim}
\end{tcolorbox}

\captionof{figure}{\textbf{Prompt D (Chinese original).}
The original Chinese YAML prompt for the compact modern-Chinese reasoning control. This control rewrites modern Chinese reasoning traces into concise modern Chinese rather than wenyan.}
\label{fig:prompt_compact_modern_zh}

\section{Ablation on the Supervised Target (Full Results)}
\label{app:ablation_train_target}

This appendix reports the full results of the ablation described in Section~\ref{sec:ablation_train_target}.
We compare the default full-completion SFT setting against a variant that masks the final answer and applies loss only to the reasoning span up to \texttt{</think>}, using the same Wenyan-rewritten training data and base model.

\begin{table}[htbp]
\centering
\small
\setlength{\tabcolsep}{5.0pt}
\resizebox{\linewidth}{!}{
\begin{tabular}{lcccc}
\toprule
Training target during SFT & 1024 & 2048 & 4096 & 6144 \\
\midrule
Full completion (\textit{thinking + </think> + answer}) & 52.9 & 55.3 & 55.4 & 55.6 \\
Thinking only (\textit{thinking + </think>}) & 52.4 & 54.1 & 54.6 & 54.9 \\
\midrule
Drop from full completion & -0.5 & -1.2 & -0.8 & -0.7 \\
\bottomrule
\end{tabular}
}
\caption{Exact-match (\%) on \textsc{MMLU-ProX-Lite-ZH}: full-completion vs.\ thinking-only supervision across generation budgets.}
\label{tab:ablation_train_target_full}
\end{table}

Removing direct supervision on the final answer causes small drops of 0.5--1.2 points across budgets. This suggests that most of the gain comes from learning a compact and stable Wenyan reasoning style, while supervising the answer channel provides a modest additional benefit by stabilizing the reasoning-to-answer transition.

\section{Case Study: Reasoning Trace Comparison on an MMLU-ProX Physics Example}
\label{app:reasoning_case_study}

\tcbset{
  querybox/.append style={
    fontupper=\footnotesize,
    fonttitle=\bfseries\footnotesize
  },
  zhbox/.append style={
    fontupper=\scriptsize,
    fonttitle=\bfseries\scriptsize
  },
  enbox/.append style={
    fontupper=\scriptsize,
    fonttitle=\bfseries\scriptsize
  }
}

We present one representative example from \texttt{mmlu\_prox\_lite\_zh} (physics, \texttt{doc\_id=10}) to qualitatively compare the reasoning traces generated by different models and prompt variants.
The gold answer is \textbf{C}, and all compared systems answer this example correctly (\texttt{exact\_match=1}).
The comparison includes five systems: (1) Qwen3.5-397B-A17B + wenyan normal prompt, (2) vernacular-reasoning SFT, (3) Qwen3-4B-Thinking-2507, (4) Qwen3.5-397B-A17B + wenyan faithful prompt, and (5) Qwen3.5-397B-A17B + wenyan ultra-compact prompt.

\subsection{Problem Statement}

\begin{tcolorbox}[querybox, title={Original Chinese Question}]
第二宇宙速度（或逃逸速度）是物体摆脱质量为 \(M\)、半径为 \(R\) 的天体引力场所需的速度。以下哪个公式正确计算了这个速度？（\(G\)：引力常量）

\vspace{0.5em}
\textbf{Options:}

A. \(v = 2\sqrt{GM/R}\) \\
B. \(v = \sqrt{GM/2R}\) \\
C. \(v = \sqrt{2GM/R}\) \\
D. \(v = \sqrt{GM \cdot R}\) \\
E. \(v = \sqrt{GM/R}\) \\
F. \(v = \sqrt{GM}\) \\
G. \(v = 2GM/R\) \\
H. \(v = \sqrt{GM/2}\) \\
I. \(v = 2\sqrt{2GM/R}\) \\
J. \(v = \sqrt{2GM}\)

\vspace{0.5em}
\textbf{Gold answer:} \(C\)
\end{tcolorbox}

\begin{tcolorbox}[querybox, title={English Translation of the Question}]
The second cosmic velocity (i.e., the escape velocity) is the speed required for an object to escape the gravitational field of a celestial body with mass \(M\) and radius \(R\).
Which of the following formulas correctly computes this speed? (\(G\): gravitational constant)

\vspace{0.5em}
\textbf{Options:}

A. \(v = 2\sqrt{GM/R}\) \\
B. \(v = \sqrt{GM/2R}\) \\
C. \(v = \sqrt{2GM/R}\) \\
D. \(v = \sqrt{GM \cdot R}\) \\
E. \(v = \sqrt{GM/R}\) \\
F. \(v = \sqrt{GM}\) \\
G. \(v = 2GM/R\) \\
H. \(v = \sqrt{GM/2}\) \\
I. \(v = 2\sqrt{2GM/R}\) \\
J. \(v = \sqrt{2GM}\)

\vspace{0.5em}
\textbf{Gold answer:} \(C\)
\end{tcolorbox}

\subsection{Chinese Reasoning Comparison}

\begin{tcolorbox}[zhbox, title={Qwen3.5-397B-A17B + wenyan normal prompt (Chinese)}]
题求第二宇宙速度，即逃逸速度。依定义，物体需具足够能量以摆脱天体引力，动能等于引力势能。设物体质量为 \(m\)，速度为 \(v\)，天体质量 \(M\)，半径 \(R\)。动能为 \((1/2)mv^2\)，引力势能为 \(-GMm/R\)。当动能等于势能绝对值时，得 \((1/2)mv^2 = GMm/R\)。消去 \(m\)，得 \((1/2)v^2 = GM/R\)。移项得 \(v^2 = 2GM/R\)。开方得 \(v = \sqrt{2GM/R}\)。比对选项，C 项为 \(\sqrt{2GM/R}\)，与推导结果一致。余项如 A 项系数为 2，D 项含 \(R\)，E 项系数为 1，F 项缺 \(R\)，G 项为速度平方，H 项系数为 \(1/2\)，I 项系数为 \(2\sqrt{2}\)，J 项缺 \(R\)，皆不符。故正确选项为 C。

\vspace{0.5em}
\textbf{Final output:} 答案是 (C)
\end{tcolorbox}

\begin{tcolorbox}[zhbox, title={Vernacular-Reasoning SFT (Chinese)}]
嗯，我现在要解决这个关于第二宇宙速度的选择题。题目是问哪个公式正确计算了逃逸速度，也就是摆脱天体引力场所需的最小速度。选项从 A 到 J，我需要仔细分析每个选项的正确性。

首先，我记得逃逸速度的公式应该和天体的质量 \(M\) 和半径 \(R\) 有关，而且应该和引力常量 \(G\) 有关。可能涉及到平方根，因为速度是标量，而动能和势能的关系可能需要平方根的形式。

那第二宇宙速度，也就是逃逸速度，正确的公式应该是怎样的呢？我记得第一宇宙速度是环绕速度，公式是 \(\sqrt{GM/R}\)，而第二宇宙速度应该是这个的 \(\sqrt{2}\) 倍，也就是 \(\sqrt{2GM/R}\)。这样的话，选项 C 应该是正确的。不过我得仔细核对一下，避免记混了。

让我再仔细推导一下。假设一个物体要脱离天体的引力，需要达到逃逸速度。根据能量守恒，物体的动能必须等于它克服引力势能所做的功。当物体在天体表面时，引力势能是 \(-GMm/R\)，动能是 \((1/2)mv^2\)。当物体逃逸时，动能全部转化为势能，所以有：
\[
\frac{1}{2}mv^2 = \frac{GMm}{R}
\]
两边都除以 \(m\)，得到：
\[
\frac{1}{2}v^2 = \frac{GM}{R}
\]
然后两边乘以 2：
\[
v^2 = \frac{2GM}{R}
\]
所以 \(v = \sqrt{2GM/R}\)，也就是选项 C。那选项 C 是正确的。

不过，我需要检查其他选项是否有可能混淆。比如选项 A 是 \(2\sqrt{GM/R}\)，这相当于第一宇宙速度的 \(\sqrt{2}\) 倍乘以 2，可能不对。选项 E 是 \(\sqrt{GM/R}\)，这其实是第一宇宙速度，所以不对。选项 C 是 \(\sqrt{2GM/R}\)，符合推导的结果。选项 I 是 \(2\sqrt{2GM/R}\)，这可能太大了，不符合能量守恒的推导。选项 J 是 \(\sqrt{2GM}\)，这没有考虑半径 \(R\)，显然不对。选项 G 是 \(2GM/R\)，单位不对。选项 D 是 \(\sqrt{GM \cdot R}\)，单位也不对。选项 H 是 \(\sqrt{GM/2}\)，系数不对。选项 B 是 \(\sqrt{GM/(2R)}\)，分母上多了 2，也不对。选项 F 是 \(\sqrt{GM}\)，同样没有考虑 \(R\)，也不对。所以正确的应该是选项 C。

\vspace{0.5em}
\textbf{Final output:} 答案是 C
\end{tcolorbox}

\begin{tcolorbox}[zhbox, title={Compact Modern Chinese Control (Chinese)}]
第二宇宙速度定义为物体摆脱天体引力场所需的最小速度，对应动能等于引力势能的临界条件。设物体质量为 \(m\)，天体质量为 \(M\)，半径为 \(R\)，逃逸时动能为 0。由能量守恒：
\[
\frac{1}{2}mv^2 = \frac{GMm}{R}
\]
消去 \(m\) 得
\[
v^2 = \frac{2GM}{R}
\]
开方解得
\[
v = \sqrt{2GM/R}
\]
对比选项，C 项公式与推导结果一致。

\vspace{0.5em}
\textbf{Final output:} 答案是 (C)
\end{tcolorbox}

\begin{tcolorbox}[zhbox, title={Qwen3-4B-Thinking-2507 (Chinese)}]
首先，问题是关于第二宇宙速度，也就是逃逸速度。逃逸速度是物体摆脱天体引力场所需的最小速度。天体的质量是 \(M\)，半径是 \(R\)，\(G\) 是引力常量。

我需要回忆逃逸速度的公式。我记得逃逸速度的公式是 \(v=\sqrt{2GM/R}\)。让我确认一下。

在物理学中，逃逸速度是从天体表面逃逸所需的最小速度。当物体以这个速度逃逸时，它的动能等于它克服引力势能所需的能量。引力势能是 \(-GMm/R\)，其中 \(m\) 是物体的质量。动能是 \((1/2)mv^2\)。

为了逃逸，初始总能量必须为 0：
\[
\frac{1}{2}mv^2 - \frac{GMm}{R} = 0
\]
所以：
\[
\frac{1}{2}mv^2 = \frac{GMm}{R}
\]
两边同时除以 \(m\)：
\[
\frac{1}{2}v^2 = \frac{GM}{R}
\]
于是：
\[
v^2 = \frac{2GM}{R}
\]
因此：
\[
v=\sqrt{\frac{2GM}{R}}.
\]

现在看选项，C 是 \(v=\sqrt{2GM/R}\)，与推导一致。A 是 \(2\sqrt{GM/R}\)，不对；B 是 \(\sqrt{GM/2R}\)，不对；D 是 \(\sqrt{GM\cdot R}\)，不对；E 是 \(\sqrt{GM/R}\)，这是第一宇宙速度；F、H、J 都缺少 \(R\)；G 的量纲不对；I 则过大。因此只有 C 正确。

\vspace{0.5em}
\textbf{Final output:} 答案是 (C)
\end{tcolorbox}

\begin{tcolorbox}[zhbox, title={Qwen3.5-397B-A17B + wenyan faithful prompt (Chinese)}]
今解第二宇宙速度之题。题云第二宇宙速度乃物体摆脱质量为 \(M\)、半径为 \(R\) 之天体引力场所需之速，\(G\) 为引力常量。当求正确公式。

首忆第二宇宙速度之义，即逃逸速度。凡逃逸速度，皆指物体脱离天体引力场所需之最小初速度。此速须足使物体克服天体引力，直至无穷远处动能为零。

次思其公式。逃逸速度之公式，当与天体质量 \(M\) 及半径 \(R\) 有关。引力势能公式为 \(-GMm/r\)，其中 \(m\) 为物体质量，\(r\) 为距天心之距离。当物体达逃逸速度时，其动能应等于克服引力势能所需之功。

设物体质量为 \(m\)，初速为 \(v\)，初时在天体表面，即 \(r=R\) 处。此时动能为 \((1/2)mv^2\)，引力势能为 \(-GMm/R\)。当物体逃逸至无穷远时，动能与势能皆为零。依能量守恒定律，初态总能量等于末态总能量：
\[
\frac{1}{2}mv^2 - \frac{GMm}{R} = 0
\]
解此方程以求 \(v\)。两边同除 \(m\)，得：
\[
\frac{1}{2}v^2 - \frac{GM}{R} = 0
\]
移项得：
\[
\frac{1}{2}v^2 = \frac{GM}{R}
\]
两边乘 2：
\[
v^2 = \frac{2GM}{R}
\]
开方得：
\[
v = \sqrt{2GM/R}
\]

观选项，C 项为 \(\sqrt{2GM/R}\)，似即正确答案。然需复核，以防有误。再查推导过程。能量守恒无误，末态动能与势能皆为零，故初态总能量应为零。是故上式确然无误，故选项 C 正确。

复观他项，如 A 项为 \(2\sqrt{GM/R}\)，此或为第一宇宙速度之二倍？第一宇宙速度乃环绕速度，公式为 \(\sqrt{GM/R}\)，故 A 项非也。又如 I 项为 \(2\sqrt{2GM/R}\)，此于本题条件下亦非也。综上，正确选项当为 C。

\vspace{0.5em}
\textbf{Final output:} 答案是 (C)
\end{tcolorbox}

\begin{tcolorbox}[zhbox, title={Qwen3.5-397B-A17B + wenyan ultra-compact prompt (Chinese)}]
第二宇宙速度乃逃逸速度。由能量守恒，动能等于引力势能绝对值：
\[
\frac{1}{2}mv^2 = \frac{GMm}{R}
\]
消 \(m\)，得
\[
v^2 = \frac{2GM}{R}
\]
开方，
\[
v=\sqrt{2GM/R}
\]
核选项，惟 C 项符此式。A 项系数为 2，非平方根；B、H、I 项因子配置不当；D、E、F、G、J 项结构不符。故 C 正确。

\vspace{0.5em}
\textbf{Final output:} 答案是 (C)
\end{tcolorbox}

\subsection{English Translation Comparison}

The following translations are provided for readability. The original model outputs were generated in Chinese.

\begin{tcolorbox}[enbox, title={Qwen3.5-397B-A17B + wenyan normal prompt (English Translation)}]
The problem asks for the second cosmic velocity, namely the escape velocity. By definition, an object must have enough energy to escape the gravitational pull of the celestial body, so its kinetic energy must equal the magnitude of the gravitational potential energy. Let the object's mass be \(m\), its speed be \(v\), the celestial body's mass be \(M\), and its radius be \(R\). The kinetic energy is \((1/2)mv^2\), and the gravitational potential energy is \(-GMm/R\). When the kinetic energy equals the absolute value of the potential energy, we have \((1/2)mv^2 = GMm/R\). Cancel \(m\), obtaining \((1/2)v^2 = GM/R\). Rearranging gives \(v^2 = 2GM/R\). Taking the square root yields \(v = \sqrt{2GM/R}\). Comparing the options, option C is \(\sqrt{2GM/R}\), which matches the derivation. The other options do not match: for instance, A has coefficient 2, D contains \(R\) in the wrong place, E has coefficient 1, F lacks \(R\), G is the squared velocity rather than velocity, H has coefficient \(1/2\), I has coefficient \(2\sqrt{2}\), and J lacks \(R\). Therefore, the correct option is C.

\vspace{0.5em}
\textbf{Final output:} The answer is (C).
\end{tcolorbox}

\begin{tcolorbox}[enbox, title={Vernacular-Reasoning SFT (English Translation)}]
Hmm, now I need to solve this multiple-choice question about the second cosmic velocity. The question asks which formula correctly computes the escape velocity, that is, the minimum speed needed to escape the gravitational field of a celestial body. The options range from A to J, so I need to carefully analyze each one.

First, I remember that the escape-velocity formula should depend on the celestial body's mass \(M\), radius \(R\), and the gravitational constant \(G\). It probably involves a square root, because velocity is a scalar quantity and the relation between kinetic energy and potential energy usually leads to a square-root form.

So what exactly is the correct formula for the second cosmic velocity, i.e., the escape velocity? I remember that the first cosmic velocity is the orbital velocity, \(\sqrt{GM/R}\), and the second cosmic velocity should be \(\sqrt{2}\) times that, namely \(\sqrt{2GM/R}\). If so, option C should be correct. Still, I should verify it carefully to avoid mixing things up.

Let me derive it more carefully. Suppose an object needs to escape the gravity of the celestial body, so it must reach the escape velocity. By energy conservation, the object's kinetic energy must equal the work required to overcome the gravitational potential. At the surface of the celestial body, the gravitational potential energy is \(-GMm/R\), and the kinetic energy is \((1/2)mv^2\). When the object escapes, all of the kinetic energy is converted into potential energy, so we have
\[
\frac{1}{2}mv^2 = \frac{GMm}{R}.
\]
Dividing both sides by \(m\), we get
\[
\frac{1}{2}v^2 = \frac{GM}{R}.
\]
Then multiplying both sides by 2:
\[
v^2 = \frac{2GM}{R}.
\]
So \(v = \sqrt{2GM/R}\), which is option C. Therefore, option C is correct.

Still, I should check whether the other options might be confusing. For example, option A is \(2\sqrt{GM/R}\), which is not right. Option E is \(\sqrt{GM/R}\), which is actually the first cosmic velocity, so it is not correct. Option I is \(2\sqrt{2GM/R}\), which is too large. Option J is \(\sqrt{2GM}\), which ignores the radius \(R\). Option G is \(2GM/R\), whose units are incorrect. Option D is \(\sqrt{GM\cdot R}\), whose units are also incorrect. Option H has the wrong coefficient. Option B has an extra 2 in the denominator. Option F also ignores \(R\). So the correct option should indeed be C.

\vspace{0.5em}
\textbf{Final output:} The answer is C.
\end{tcolorbox}

\begin{tcolorbox}[enbox, title={Compact Modern Chinese Control (English Translation)}]
The second cosmic velocity is defined as the minimum speed required for an object to escape the gravitational field of a celestial body. It corresponds to the critical condition where the kinetic energy equals the gravitational potential energy. Let the object's mass be \(m\), the celestial body's mass be \(M\), and its radius be \(R\). At escape, the final kinetic energy is zero. By energy conservation,
\[
\frac{1}{2}mv^2 = \frac{GMm}{R}.
\]
Canceling \(m\), we obtain
\[
v^2 = \frac{2GM}{R}.
\]
Taking the square root gives
\[
v = \sqrt{2GM/R}.
\]
Comparing the options, option C matches the derived formula.

\vspace{0.5em}
\textbf{Final output:} The answer is (C).
\end{tcolorbox}

\begin{tcolorbox}[enbox, title={Qwen3-4B-Thinking-2507 (English Translation)}]
First, the problem is about the second cosmic velocity, namely the escape velocity. The escape velocity is the minimum speed required for an object to escape the gravitational field of a celestial body. The celestial body's mass is \(M\), its radius is \(R\), and \(G\) is the gravitational constant.

I need to recall the formula for escape velocity. I remember that the formula is \(v=\sqrt{2GM/R}\). Let me verify that.

In physics, the escape velocity is the minimum speed required to escape from the surface of a celestial body. When an object escapes at this speed, its kinetic energy equals the energy required to overcome the gravitational potential. The gravitational potential energy is \(-GMm/R\), where \(m\) is the object's mass, and the kinetic energy is \((1/2)mv^2\).

To escape, the initial total energy must be zero:
\[
\frac{1}{2}mv^2 - \frac{GMm}{R} = 0.
\]
So,
\[
\frac{1}{2}mv^2 = \frac{GMm}{R}.
\]
Dividing both sides by \(m\),
\[
\frac{1}{2}v^2 = \frac{GM}{R}.
\]
Thus,
\[
v^2 = \frac{2GM}{R}.
\]
Therefore,
\[
v=\sqrt{\frac{2GM}{R}}.
\]

Now compare the options: C is \(v=\sqrt{2GM/R}\), which matches the derivation. A is \(2\sqrt{GM/R}\), which is wrong; B is \(\sqrt{GM/2R}\), which is wrong; D is \(\sqrt{GM\cdot R}\), which is wrong; E is \(\sqrt{GM/R}\), which is the first cosmic velocity; F, H, and J all omit \(R\); G has the wrong dimensional form; and I is too large. Therefore, only C is correct.

\vspace{0.5em}
\textbf{Final output:} The answer is (C).
\end{tcolorbox}

\begin{tcolorbox}[enbox, title={Qwen3.5-397B-A17B + wenyan faithful prompt (English Translation)}]
Now let us solve the problem of the second cosmic velocity. The question states that the second cosmic velocity is the speed required for an object to escape the gravitational field of a celestial body with mass \(M\) and radius \(R\), where \(G\) is the gravitational constant. We are to determine the correct formula.

First, recall the meaning of the second cosmic velocity: it is the escape velocity. In general, the escape velocity refers to the minimum initial speed required for an object to leave the gravitational field of a celestial body. This speed must be sufficient for the object to overcome the body's gravity until its kinetic energy becomes zero at infinity.

Next, consider the formula. The escape-velocity formula should depend on the celestial body's mass \(M\) and radius \(R\). The gravitational potential-energy formula is \(-GMm/r\), where \(m\) is the mass of the object and \(r\) is the distance from the body's center. When the object reaches the escape velocity, its kinetic energy should equal the work required to overcome the gravitational potential energy.

Let the object's mass be \(m\), its initial speed be \(v\), and let it start from the body's surface, i.e., \(r=R\). Then its kinetic energy is \((1/2)mv^2\), and its gravitational potential energy is \(-GMm/R\). When the object escapes to infinity, both the kinetic energy and the potential energy become zero. By conservation of energy, the total energy in the initial state equals the total energy in the final state:
\[
\frac{1}{2}mv^2 - \frac{GMm}{R} = 0.
\]
Solving for \(v\), divide both sides by \(m\):
\[
\frac{1}{2}v^2 - \frac{GM}{R} = 0.
\]
Rearranging:
\[
\frac{1}{2}v^2 = \frac{GM}{R}.
\]
Multiplying both sides by 2:
\[
v^2 = \frac{2GM}{R}.
\]
Taking the square root:
\[
v=\sqrt{2GM/R}.
\]

Looking at the options, option C is \(\sqrt{2GM/R}\), which seems to be the correct answer. Still, one should verify again to avoid error. Re-checking the derivation, energy conservation is indeed applied correctly: both kinetic energy and potential energy are zero in the final state, so the initial total energy must also be zero. Hence option C is correct.

Looking again at the other options, option A is \(2\sqrt{GM/R}\), which could be taken as twice the first cosmic velocity. Yet the first cosmic velocity is the orbital velocity, \(\sqrt{GM/R}\), so A is not correct. Likewise, option I is \(2\sqrt{2GM/R}\), which does not fit the conditions of this problem either. In sum, the correct option should be C.

\vspace{0.5em}
\textbf{Final output:} The answer is (C).
\end{tcolorbox}

\begin{tcolorbox}[enbox, title={Qwen3.5-397B-A17B + wenyan ultra-compact prompt (English Translation)}]
The second cosmic velocity is the escape velocity. By energy conservation, the kinetic energy equals the absolute value of the gravitational potential energy:
\[
\frac{1}{2}mv^2 = \frac{GMm}{R}.
\]
Cancel \(m\), obtaining
\[
v^2 = \frac{2GM}{R}.
\]
Taking the square root,
\[
v=\sqrt{2GM/R}.
\]
Checking the options, only C matches this formula. In A the coefficient is 2 rather than being under the square root; B, H, and I contain mismatched factors; D, E, F, G, and J have the wrong structure. Therefore, C is correct.

\vspace{0.5em}
\textbf{Final output:} The answer is (C).
\end{tcolorbox}

\subsection{Brief Observation}

This example illustrates the qualitative differences among long-form thinking, compact modern-Chinese reasoning, and wenyan reasoning. The \textbf{vernacular-reasoning SFT} and \textbf{Qwen3-4B-Thinking-2507} outputs are substantially more verbose and conversational, containing more discourse management, self-verification, and option-by-option checking. The \textbf{Compact Modern Chinese control} removes most of this colloquial and repetitive content while preserving the essential derivation in standard modern Chinese. It therefore serves as a useful control showing that generic shortening alone can already produce a much denser reasoning trace.

The wenyan variants further change the linguistic register of the reasoning channel. The \textbf{wenyan normal prompt} yields a short and dense trace while retaining the main derivation and lightweight option filtering. The \textbf{wenyan faithful prompt} preserves more step-by-step structure and verification behavior, but expresses it in a more classical style. The \textbf{wenyan ultra-compact prompt} produces the most skeletal trace, retaining only the minimal derivation and final option check. Overall, the case study supports our interpretation that compact rewriting is an important factor, while wenyan provides a distinct compact reasoning register rather than merely a shorter version of vernacular reasoning.

\end{document}